\documentclass[journal]{IEEEtran}
\usepackage{times}
\usepackage{epsfig}
\usepackage{graphicx}
\usepackage{amsmath,amsfonts}
\usepackage{amssymb}
\usepackage{array}
\usepackage{multirow}
\usepackage{subcaption}
\usepackage{color}
\usepackage{float}
\usepackage{graphicx}
\usepackage{mathtools}
\usepackage{booktabs}
\usepackage{hyperref}
\usepackage{mathtools}
\usepackage{dblfloatfix}
\usepackage{cite}
\usepackage{graphics}

\newcases{Rcases}{$\mskip\thickmuskip$}{%
  \hfil$\m@th{##}$}{$\m@th{##}$\hfil}{.}{\rbrace}

\usepackage[font=scriptsize]{caption}  
\usepackage[normalem]{ulem}
\usepackage{xcolor}
\definecolor{darkgreen}{rgb}{0,.4,0}
\definecolor{darkcyan}{rgb}{0,.4,.4}
\newcommand{\REMOVE}[1]%
          {{\color{blue}\sout{#1}}}

\newcommand{\COMMENT}[1]%
          {{\color{darkgreen}\textbf{{Editor: }} {#1}}}

\begin{document}


\title{Neighborhood Attention Makes the Encoder of ResUNet Stronger for Accurate Road Extraction}

\author{
 
 Ali Jamali,
 Swalpa Kumar Roy,~\IEEEmembership{Student Member,~IEEE,} \\
 Jonathan Li,~\IEEEmembership{Fellow Member,~IEEE,}
 and Pedram Ghamisi,~\IEEEmembership{Senior Member,~IEEE}
\thanks{This research was funded by the Institute of Advanced Research in Artificial Intelligence (IARAI).(Corresponding author: \textit{Pedram Ghamisi})}
\thanks{A. Jamali is with the Department of Geography, Simon Fraser University, British Columbia 8888, Canada (e-mail: alij@sfu.ca).}
\thanks{S. K. Roy is with the Department of Computer Science and Engineering, Jalpaiguri Government Engineering College, West Bengal 735102, India (e-mail: swalpa@cse.jgec.ac.in).}
\thanks{ J. Li is with the University of Waterloo Department of Geography and Environmental Management, Waterloo, Ontario, Canada (e-mail: junli@uwaterloo.ca).}
\thanks{P. Ghamisi is with the Helmholtz-Zentrum Dresden-Rossendorf (HZDR), Helmholtz Institute Freiberg for Resource Technology, 09599 Freiberg, Germany, and is also with the Institute of Advanced Research in Artificial Intelligence (IARAI), 1030 Vienna, Austria (e-mail: p.ghamisi@gmail.com).}
}

\markboth{SUBMITTED TO IEEE JOURNAL}
 {Jamali \MakeLowercase{\textit{et al.}}: Bare Demo of IEEEtran.cls for Journals}

\maketitle

\begin{abstract}
In the domain of remote sensing image interpretation, road extraction from high-resolution aerial imagery has already been a hot research topic. Although deep CNNs have presented excellent results for semantic segmentation, the efficiency and capabilities of vision transformers are yet to be fully researched. As such, for accurate road extraction, a deep semantic segmentation neural network that utilizes the abilities of residual learning, HetConvs, UNet, and vision transformers, which is called \texttt{ResUNetFormer}, is proposed in this letter. The developed \texttt{ResUNetFormer} is evaluated on various cutting-edge deep learning-based road extraction techniques on the public Massachusetts road dataset. Statistical and visual results demonstrate the superiority of the \texttt{ResUNetFormer} over the state-of-the-art CNNs and vision transformers for segmentation. The code will be made available publicly at \url{https://github.com/aj1365/ResUNetFormer}
\end{abstract}

\begin{IEEEkeywords}
Vision transformers, road extraction, attention mechanism, UNet, neighbourhood attention transformer~(NAT).
\end{IEEEkeywords}

\IEEEpeerreviewmaketitle

\section{Introduction}

\IEEEPARstart{O}{ne} of the most profound tasks in the area of remote sensing is the accurate road extraction. Despite substantial interest in the last decade, road extraction from high-resolution imagery remains difficult due to occlusions, noise, and the difficulty of the surrounding features in remotely sensed imagery \cite{8309343}. Deep neural network-based techniques have achieved a high level of performance on a broad range of tasks in computer vision, including road detection \cite{8792386, 8883072}. Recently, due to the great success achieved by vision transformers (ViTs) in Natural Language Processing (NLP), scientists in the remote sensing community have also been attempting to harness the power of vision transformers to address the issues of interpretation of complex and high-dimensional remote sensing data \cite{9627165}. These methods have proven superior performance to conventional and CNN-based models \cite{jamali2022deep}, demonstrating the immense promise of using such methods to understand and analyze remote sensing imagery.

Nevertheless, due to issues such as vanishing gradients, training a very deep architecture is incredibly challenging \cite{8309343}. To address this issue, He \textit{et al.} \cite{he2016deep} proposed the deep residual learning method, which employs identity mapping to aid in training. Ronneberger \textit{et al.} \cite{ronneberger2015u} introduced the UNet, which concatenates feature maps from various levels to enhance segmentation performance, rather than using skip connections in fully convolutional networks (FCNs) \cite{long2015fully}. UNet incorporates low-level detailed information with high-level semantic representation, resulting in promising biomedical image segmentation performance \cite{ronneberger2015u}. On the other hand, ViTs utilize self attention mechanism rather than the widely used convolutional operations employed by standard deep models \cite{dosovitskiy2020image}. Consequently, unlike CNNs, ViTs capture global contextual information in a better way with the utilization of self-attention at the cost of quadratic complexity. This helps the transformers to outperform the CNN algorithms in terms of feature generalization capabilities. Moreover, because of their flexible attention window, ViTs, such as the neighborhood attention transformer (NAT) \cite{Hassani2022}, have demonstrated the potential of linear computational costs and gains more attention in the vision community. As such, we propose the ResUNetFormer that integrates the capabilities of  heterogeneous convolution (HetConv), residual learning, UNet, and NAT for accurate prediction of road information from high-resolution aerial imagery. We developed a deep learning U-Net based semantic segmentation framework that effectively utilizes HetConv operation to leverage heterogeneous kernels within the residual learning unit for degradation free feature representation learning. In contrast with the conventional vanilla ViTs, the proposed model utilizes NAT, which replaces the computationally expensive self-attention mechanism for enhancing the feature generalization ability limited within a local neighborhood that substantially reduces the computation cost. The use of residual learning and UNet with NAT resulted in much lower segmentation noises as compared to the cutting-edge vision transformers-based segmentation models, i.e., SwinUNet \cite{cao2021swin} and Attention UNet \cite{8309343}. 

This letter introduces the ResUNetFormer in Section~\ref{sec:prop}, presents the experiments and analysis in Section~\ref{sec:exp}, and highlights the concluding remarks in Section~\ref{sec:con}.


\begin{figure*}
\centering
\includegraphics[clip=true, trim= 02 300 02 15, width=0.78\textwidth]{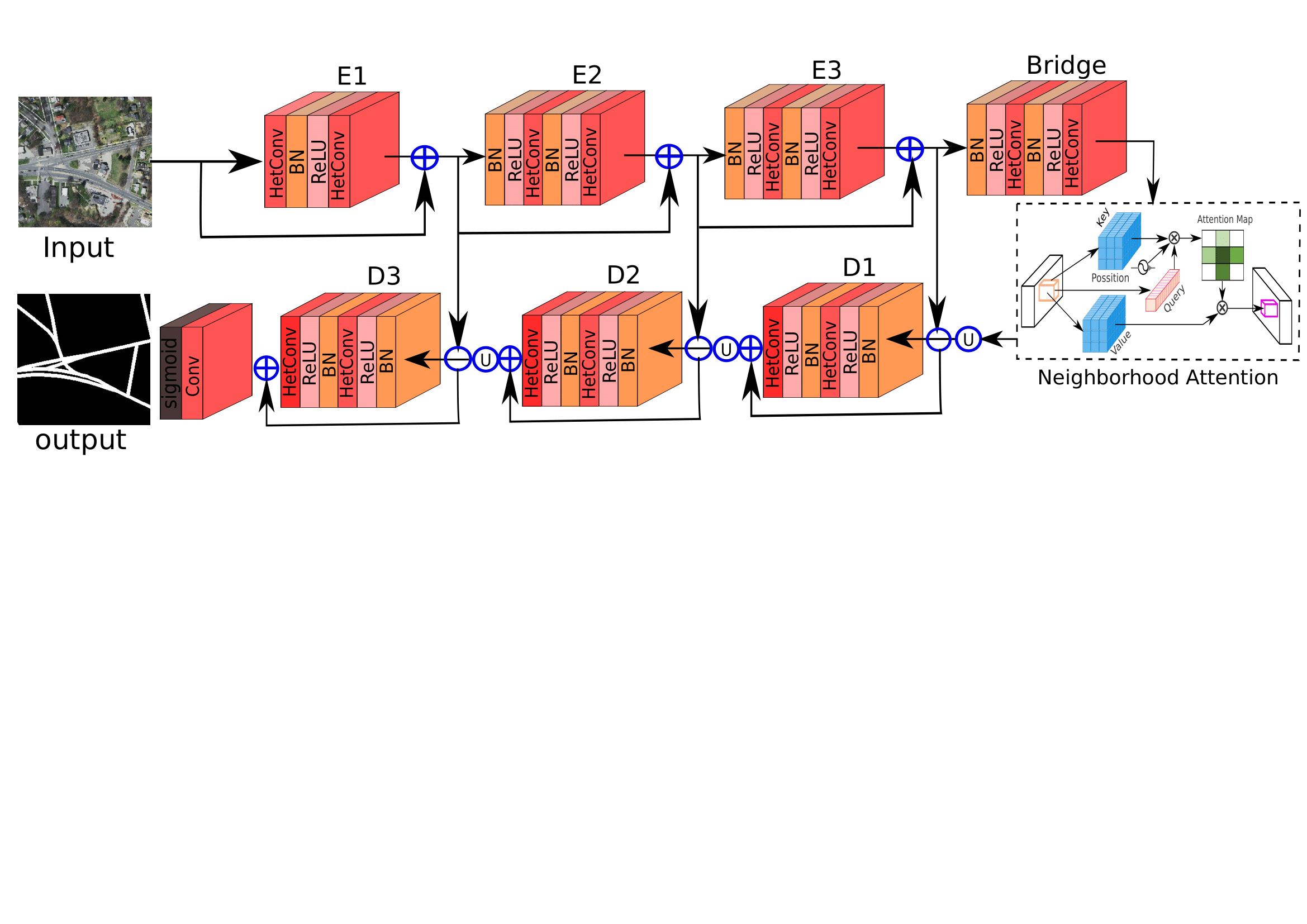}
\caption{Proposed ResUNetFormer model for accurate extraction of road information where $\oplus$ and $\ominus$ represent element wise addition and concatenation operations and circular $U$ denotes the up-sampling operation of the feature maps.}
\label{fig:attention}
\end{figure*}

\section{Proposed segmentation Framework}
\label{sec:prop}
Given an image $\textbf{X}\in R^{H\times{W}\times{C}}$ where $H$ and $W$ represent spatial height and width and $C$ is the number of channels, respectively. The goal is to predict $y = \mathcal{F}(X)$ that corresponds to the pixel label classification map of input $\textbf{X}$ having size of $(H\times{W})$. The easiest way to perform such a task is to use a convolutional U-network that maps input images into high-level feature representations in encoding stages, and then decode back to the full spatial resolution to produce a pixel-wise label map. In this paper, we introduce the ResUNetFormer model for semantic segmentation of the road extraction task that incorporates the advantages of HetConv, U-Network, residual skip connection, and vision transformers~(ViT). The ResUNetFormer provides several advantages: (1) HetConv combines group-wise convolution and point-wise convolution to increase the efficiency of the generalized representation; (2) the residual learning will enable effective network training; (3) the skip connections in a residual unit, which connects the low level feature to its corresponding high level feature, improve information propagation without degradation, enabling us to construct lower complexity framework that captures better semantic segmentation information with the limited amount of reference data; (4) NAT allows a pixel-level operation that localizes each pixel to its neighborhood and improves the acquisition of global and local contextual information from satellite imagery.

\noindent 1) \textbf{Residual learning:} To increase the projection power of CNNs, one of the traditional ways is to add more layers to the network, which may hampers accurate propagation of models information during back-propagation and yields degradation gradient \cite{he2016deep}. To tackle these shortfalls, a residual unit was introduced in the place of the conventional convolutional blocks, which carefully monitor the vanishing gradient issue with a skip connection and residual learning. The skip connection of a residual unit allows to map low level feature representation to its high level representation in an easier way. Suppose $\mathbf{X_j^{l-1}}$ represents the input $j$th feature map of the residual blocks which is parameterized with $F_{RN}(\mathbf{X}^{l-1};\theta_{1}, \theta_{2})$ using HetConv filter banks $W = \{W^{l+i}|1\leq i\leq 2\}$ of kernel size  $k_{1}^{l}\times{k_{2}^{l}}$ in the ${(l-1)}$th and ${(l)}^{th}$ layers, respectively. The output feature map $X^{l+1}$ obtained in the $(l+1)$th layer can be calculated as follows:
\begin{equation}
    {X}^{l+1} = I({X^{l-1}}) + F_{RN}({X}^{l-1};{\theta_1}, {\theta_2})
    \label{equ:res} 
\end{equation}
\noindent where $I({X^{l-1}})$ the identity mapping in a residual unit.
\begin{equation}
    \begin{aligned}
      F_{RN}(X^{l-1};\theta_{1}, \theta_{2}) = ReLU(BN(X^{l}\circledast W^{l+1} + b^{l+1}) \\
      X^{l} = ReLU(BN(X^{l-1}\circledast W^{l} + b^{l}))  
    \end{aligned}
\end{equation}


\noindent where $\circledast$ represents the \textit{HetConv} operation and $X^{l}$ and $X^{l+1}$ are output feature maps in $l$th and ${(l+1)}$th layers, respectively. $\theta_1$ and $\theta_2$ denote the weight and bias parameters associated with the $j$th unit of the $l$th and ${(l+1)}$th of \textit{HetConv} layers, respectively.

\noindent 2) \textbf{Neighborhood attention transformer (NAT):} NAT limits the receptive field of each query token to fixed-sized neighboring pixels in its local area of the neighborhood. The NAT is driven by the goal of creating a local neighborhood region in which the smaller neighboring area receives more local attention, and the wider neighboring area obtains more global attention. The local neighborhood region of all points $\forall_y$ such that $dist(x, y) \leq r$ where $r$ shows the radius of the local window, which can be expressed as $\lambda{(i,j)} = \{ y\in x: dist(x,y) \leq r \}$.

The NAT consider a pixel at location $(i,j)$ of the feature map extracted from the bridge layer of encoder is expressed as the linear projections $\Phi$ of the nput features $X\in R^{(48\times{48}\times{512})}$, the {\color{black} queries} {$Q=\Phi_q{X}$} whereas {\color{black} keys} {$K=\Phi_k{X}$} and the {\color{black} values} {$V=\Phi_v{X}$} functions for all the $ith$ input patch with local wind of size $r$ and the size of the patch matrix is expressed as $48\times{48}$, whereas $512$ is the embedding dimension of the matrix of the input feature, and $\Phi_q$, $\Phi_v$, and $\Phi_k$ represent the parameters of the projections i.e., query, value, and  key, respectively. The $\Phi$ will be optimized utilizing the Adam \cite{kingma2014adam} optimizer in the phase of model's training. To produce weights of the attention $\mathbf{A_i^k}$, the scaled dot product is utilized whiting $i$th input queries ($\mathbf{Q}$) with ($r\times{r}$) neighboring keys ($\mathbf{K}$) as follows:
\begin{equation}
A_i =
\begin{bmatrix}
 Q_{i}K_{\lambda{(i,1)}}^T + B_{\lambda{(i,1)}} \\
 Q_{i}K_{\lambda{(i,2)}}^T + B_{\lambda{(i,2)}} \\ \vdots \\
 Q_{i}K_{\lambda{(i,r^2)}}^T + B_{\lambda{(i,r^2)}} 
\end{bmatrix}
\label{equ:atten_weight}
\end{equation}

\noindent where $\lambda{(i,j)}$ defines the $j$th neighboring region of the $i$th query which is dependent on their relative positions, the corresponding location bias is expressed by $B_{\lambda{i,j}}$, which is added to each point of attention weights. Afterwards, $V_{\lambda{(i,j)}} = [ V_{\lambda{(i,1)}}, V_{\lambda{(i,2)}}, \ldots V_{\lambda{(i,r^2)}}]$ are presented by the $r^2$ localized region with the values of the $i$th input query. The output of $i$th attention weights $A_i$ in Eq.~(\ref{equ:atten_weight}) is passed through a \texttt{softmax} function to calculate the attention map. Thus, the \texttt{NAT} for the $i$th input token of the $Y_i^r$ the output map is expressed as:

\begin{equation}
    \texttt{NAT}(\mathbf{Y_i^r}) = \mathbf{\texttt{softmax}{\left( \frac{A_i^r}{\sqrt{D}} \right )}V_{\lambda{(i,j)}^r}}
\end{equation}

\noindent where the scale is defined by $1/\sqrt{D}$, which is utilized to enhance the \texttt{softmax} function's small gradient propagation. It should be noted that we used neighboring window $r = 3$ for the experimental analysis in this letter.

\noindent 3) \textbf{ResUNetFormer:} In  this  work, we adopted the 7-layer ResUNet architecture initially developed by Zhang \textit{et al.} \cite{8309343} to address the accurate road extraction task as the backbone, illustrated in Fig.~\ref{fig:attention}. The ResUNetFormer is build with four components which includes encoding, bridge, NAT, and decoding. In the encoding stage, i.e., $E1$ to $E3$, the input images are compressed into compact representations which is achieved through the consecutive use of the residual unit as shown in Eq.~(\ref{equ:res}). The final section, i.e., $D1$ to $D3$, is responsible for restoring the representations to perform pixel-by-pixel classification, i.e., semantic segmentation. The bridge section acts as a link between the encoding and NAT sections and helps the decoder for smooth recovering of the features. The input of the NAT block is the feature map of the bridge section, and the resulting attention maps will be passed into the decoder section. The encoder, bridge, and decoder are constructed by residual units, which have two $3\times{3}$ HetConv blocks and an identity mapping (defined by Eq.~(\ref{equ:res})) followed by a Batch Normalization layer, a ReLU activation layer, and a HetConv layer in each block of convolutions. It should be noted that the identity mapping connects the low-level input feature and high-level output feature in a residual unit.

The encoding part has three units of the residual function. It should be mentioned that rather than employing a pooling operation to reduce the size of output maps, a stride of $2$ was utilized in the first block of HetConvs in each unit, decreasing the ratios of output maps by 50\%. Similarly, the decoding part has three residual units. Before each unit, output maps from the lower level will be up-sampled and concatenated with their relative encoding path output maps. A 2D convolution with kernel size $(1\times{1})$ and a sigmoid activation function are employed after the last level of decoding for projecting the multichannel output maps into the targeted segmentation road map. It should be noted that we developed two versions of ResUNetFormer. In ResUNetFormer-V1, similar to the ResUNet model, we utilized Conv2D operations, while in ResUNetFormer-V2, instead of using Conv2D functions, we employed HetConv as seen in Fig.~\ref{fig:attention}. In the HetConv layers, there are three depth-wise convolutional groups with kernel sizes of $3\times{3}$, while there is a point-wise Conv2D with a kernel size of $1\times{1}$. The feature map of depth-wise convolutions is added to the point-wise convolution to produce the results of HetConv functions. The details of parameters and size of feature maps for the ResUNetFormer-V1 and ResUNetFormer-V2 are illustrated in Table~\ref{tab:parameters}.

\begin{table}[!ht]
\centering
\caption{The layer-wise architecture of the ResUNetFormer-V1 and ResUNetFormer-V2 segmentation algorithms.}
\resizebox{0.98\linewidth}{!}{
\begin{tabular}{@{}|c|c|c|c|c|c|@{}}
\toprule
 Unit level        & Filter-V1 & Filter-V2 & Stride & Output size-V1 &  Output size-V2\\ \midrule
Input &  -  &  -&  -& 384 * 384 * 3 & 384 * 384 * 3 \\ \midrule
  E1 	& 3 * 3 * 64& 	[3*(3 * 3*22)]+ [1*(1*1* 66)]& 	1	& 384 * 384 * 64& 	384 * 384 * 66
 \\ \cmidrule(l){2-6} 
   &	3 * 3 * 64&	[3*(3 * 3*22)]+ [1*(1*1* 66)]&	1	&384 * 384 * 64	&384 * 384 * 66
 \\ \midrule
 \multirow{1}{*}{} E2 & 	3 * 3 * 128&	[3*(3 * 3*42)]+ [1*(1*1* 126)]	&2	&192 * 192 * 128&	192 * 192 * 126
 \\  \cmidrule(l){2-6} 
  & 3 * 3 * 128&	[3*(3 * 3*42)]+ [1*(1*1* 126)]	&1	&192 * 192 * 128&	192 * 192 * 126
 \\ \midrule
  \multirow{1}{*}{} E3 &  	3 * 3 * 256	&[3*(3 * 3*84)]+ [1*(1*1* 252)]&	2&	96 * 96 * 256&	96 * 96 * 252
 \\  \cmidrule(l){2-6} 
  &  	3 * 3 * 256 &	[3*(3 * 3*84)]+ [1*(1*1* 252)]	 &1 &	96 * 96 * 256 &	96 * 96 * 252
  \\ \midrule
  \multirow{1}{*}{} Bridge & 	3 * 3 * 512&	[3*(3 * 3*84)]+ [1*(1*1* 510)]&	2&	48 * 48 * 512&	48 * 48 * 510
 \\   \cmidrule(l){2-6} 
  &  3 * 3 * 512	 &[3*(3 * 3*84)]+ [1*(1*1* 510)] &	1	 &48 * 48 * 512 &	48 * 48 * 510
 \\ \midrule
 NAT &   -	& -	& 1	& 48 * 48 * 512& 	48 * 48 * 510
 \\ \midrule
  \multirow{1}{*}{} D1 &	3 * 3 * 256	&[3*(3 * 3*84)]+ [1*(1*1* 252)]&	1&	96 * 96 * 256&	96 * 96 * 252
  \\  \cmidrule(l){2-6} 
  &   3 * 3 * 256	  & [3*(3 * 3*84)]+ [1*(1*1* 252)]  & 	1  & 	96 * 96 * 256  & 	96 * 96 * 252
 \\ \midrule
  \multirow{1}{*}{} D2 & 	3 * 3 * 128	&[3*(3 * 3*42)]+ [1*(1*1* 126)]&	1&	192 * 192 * 128&	192 * 192 * 126
  \\ \cmidrule(l){2-6} 
  & 	3 * 3 * 128	  &[3*(3 * 3*42)]+ [1*(1*1* 126)]  &	1  &	192 * 192 * 128	  &192 * 192 * 126
  \\ \midrule
  \multirow{1}{*}{} D3 &	3 * 3 * 64	&[3*(3 * 3*22)]+ [1*(1*1* 66)]&	1	& 384 * 384 * 64	&384 * 384 * 66
 \\  \cmidrule(l){2-6} 
  & 	3 * 3 * 64 & 	[3*(3 * 3*22)]+ [1*(1*1* 66)]& 	1& 	384 * 384 * 64 & 	384 * 384 * 66
 \\ \midrule
  Output & 	1 * 1&	1 * 1 &	1	& 384 * 384 * 1	&384 * 384 * 1
  \\ \bottomrule
 
\end{tabular}}
\label{tab:parameters}
\end{table}

\section{Experimental Results}
\label{sec:exp}
The developed model, ResUNetFormer, is evaluated against several other state-of-the-art segmentation models, including UNet \cite{ronneberger2015u}, UNet++ \cite{8932614}, UNet+++ \cite{huang2020unet}, Attention UNet \cite{oktay2018attention}, SwinUNet \cite{cao2021swin}, and ResUNet \cite{8309343}, respectively.

\subsection{Experimental Data and Settings}

Mihn \textit{et al.} \cite{mnih2010learning} created the Massachusetts roads data (MRD). The road benchmark contains 1171 high-resolution images, which include 1108 images for training, 14 for validation, and 49 for testing. The images are with the size of $1500\times{1500}$ pixels and with a spatial resolution of 1.2 meters per pixel. The MRD data benchmark represents nearly 500 km$^2$ of an area that contains urban, suburban, and rural regions, as well as a diverse range of ground artifacts, such as sea, bridges, rivers, roadways, buildings, ports, schools, and vegetation. We trained the model with images of size $384\times{384}$ in this letter. It should be mentioned that throughout training, no data augmentation was used. In this letter, we utilized a learning rate, batch size and number of epoch of 0.0001, 1 and 40, respectively.

\subsection{Segmentation Results}

To validate the efficiency of the proposed ResUNetFormer for accurate road extraction, we consider the MRD dataset to create two experimental settings. In scenario 1~(MRD100), we only used $100$ images as the training data, whereas in scenario 2~(MRD800), $800$ images were utilized to train the segmentation algorithms. We employed binary cross entropy as the loss function in scenarios 1 and 2. On the other hand, we have also used intersection over union (IoU) ($IoU =\frac{\text{Area of Overlap}}{\text{Area of Union}}$) loss function in scenario 1~(MRD100IOU) with $100$ training images and scenario 2~(MRD800IOU) with $800$ training images.

\begin{table}[!htbp]
\centering
\caption{Segmentation results of the MRD100 dataset in terms of F-1 score, Precision, Recall, and Dice coefficient.}
\resizebox{0.95\linewidth}{!}{
\begin{tabular}{|c|c|c|c|c|} \toprule
Algorithm &	F-1$\times$100 \textbf{$\uparrow$} &	Precision$\times$100 $\uparrow$ &	Recall$\times$100 $\uparrow$ & Dice coefficient $\uparrow$ \\
 \midrule
UNet \cite{ronneberger2015u}&	48.53 &	82.03 &	25.91 &	0.3149\\ 
UNet++\cite{8932614}	& 53 &	79.56 &	32.52 &	0.3315\\ 
UNet+++ \cite{huang2020unet} &	49.71 &	80 &	26.34 &	0.3392\\ 
AttUNet\cite{oktay2018attention} &	54.64 &	86.64 &	27.18 &	0.3879\\
SwinUNet \cite{cao2021swin} &	44.38 &	69.42 &	29.73 &	0.3728\\  
ResUNet \cite{8309343} & 65.6 &	91.41 &	 41.3 &	0.469 \\ 
ResUNetFormer\_V1 &	63.07 &	 \textbf{95.3} &	35.79 &	0.4394\\ 
ResUNetFormer\_V2 &	 \textbf{65.82} &	88.99 &	 \textbf{45.59} &	\textbf{0.5113}\\  
\bottomrule
\end{tabular}}
\label{tab:MRD100}
\end{table}

\begin{figure*}[!htbp]
\centering
\begin{subfigure}{.22\columnwidth}
\centering
\includegraphics[clip=true, trim = 0 0 0 0, width=0.98\linewidth]{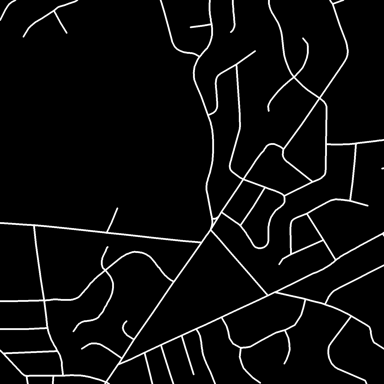} 
\caption{}
\end{subfigure}%
\begin{subfigure}{.22\columnwidth}
\centering
\includegraphics[clip=true, trim = 0 0 0 0, width=0.98\linewidth]{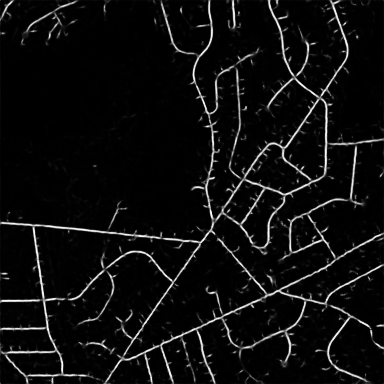} 
\caption{}
\end{subfigure}%
\begin{subfigure}{.22\columnwidth}
\centering
\includegraphics[clip=true, trim = 0 0 0 0, width=0.98\linewidth]{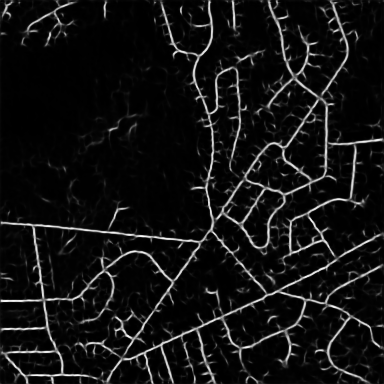} 
\caption{}
\end{subfigure}%
\begin{subfigure}{.22\columnwidth}
\centering
\includegraphics[clip=true, trim = 0 0 0 0, width=0.98\linewidth]{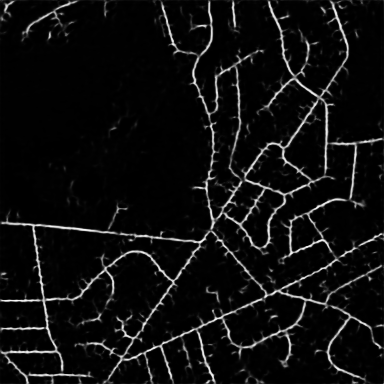} 
\caption{}
\end{subfigure}%
\begin{subfigure}{.22\columnwidth}
\centering
\includegraphics[clip=true, trim = 0 0 0 0, width=0.98\linewidth]{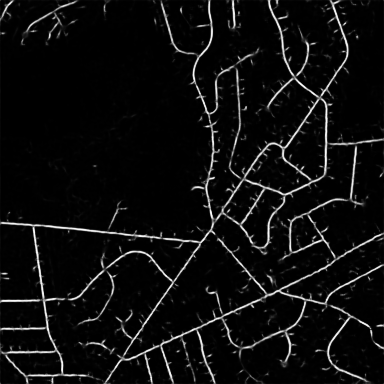} 
\caption{}
\end{subfigure}%
\begin{subfigure}{.22\columnwidth}
\centering
\includegraphics[clip=true, trim = 0 0 0 0, width=0.98\linewidth]{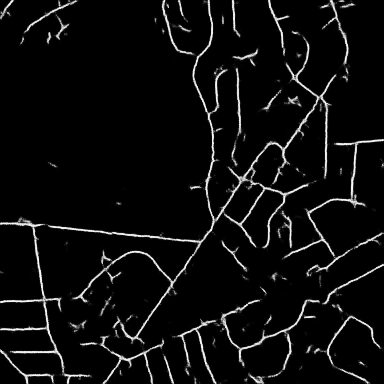} 
\caption{}
\end{subfigure}%
\begin{subfigure}{.22\columnwidth}
\centering
\includegraphics[clip=true, trim = 0 0 0 0, width=0.98\linewidth]{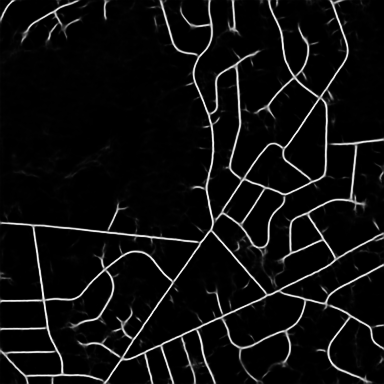} 
\caption{}
\end{subfigure}%
\begin{subfigure}{.22\columnwidth}
\centering
\includegraphics[clip=true, trim = 0 0 0 0, width=0.98\linewidth]{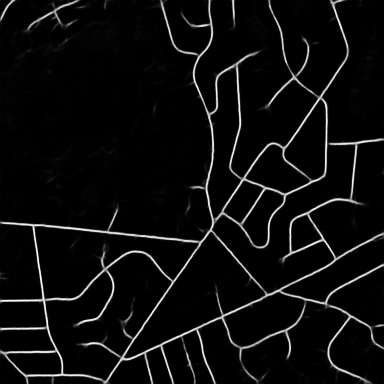} 
\caption{}
\end{subfigure}%
\begin{subfigure}{.22\columnwidth}
\centering
\includegraphics[clip=true, trim = 0 0 0 0, width=0.98\linewidth]{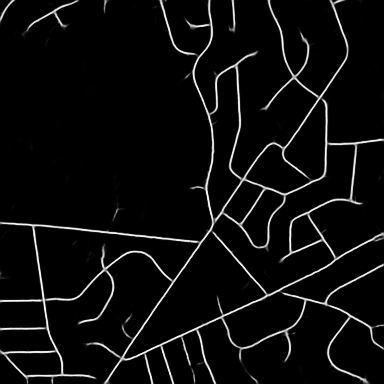} 
\caption{}
\end{subfigure}
\caption{Segmentation maps over MRD100 dataset using (a) Ground Truth, (b) UNet, (c) UNet++, (d) UNet+++, (e) Attention UNet, (f) SwinUNet, (g) ResUNet, (h) ResUNetFormer-V1, and (i) ResUNetFormer-V2.}
\label{fig:MRD100}
\end{figure*}

As seen in Table~\ref{tab:MRD100} and Fig.~\ref{fig:MRD100}, the best results in terms of recall (45.59\%), dice coefficient (0.513), F-1 score (65.82\%), and visual interpretation were achieved by the ResUNetFormer-V2 model with the HetConv operations. Moreover, the highest precision was obtained by the ResUNetFormer-V1 (95.3\%). The ResUNetFormer-V2 results increased the F-1 score, dice coefficient, and recall of the ResUNet by about 1\%, 9\%, and 10\%, respectively. In addition, the ResUNetFormer-V2 with HetConv opertions model showed significantly less noise compared to other segmentation techniques, including vision-based SwinUNet and Attention UNet.

\begin{table}[!htbp]
\centering
\caption{Segmentation results of MRD800 dataset in terms of F-1 score, Precision, Recall, and Dice coefficient.}
\resizebox{0.95\linewidth}{!}{
\begin{tabular}{|c|c|c|c|c|} \toprule
Algorithm &	F-1$\times$100 $\uparrow$ & Precision$\times$100 $\uparrow$ &Recall$\times$100 $\uparrow$ & Dice coefficient $\uparrow$ \\
 \midrule
UNet \cite{ronneberger2015u} &	51.38 &	87.96 &	26.64 &	0.4671\\ 
UNet++\cite{8932614}	& 49.74 &	89.89 &	23.19 &	0.4584\\ 
UNet+++ \cite{huang2020unet} &	51.91 &	87.15 &	28.29 &	0.4811\\ 
AttUNet\cite{oktay2018attention} &	61.56 &	92.53 &	32.08 &	0.4608\\
SwinUNet \cite{cao2021swin} &	61.52 &	86.14 &	43	 & 0.5171\\  
ResUNet \cite{8309343} & 57.2&	92.45 &	31.94 &	\textbf{0.5706} \\ 
ResUNetFormer\_V1 &	64.65 &	\textbf{98.13} &	34.1 &	0.5143\\ 
ResUNetFormer\_V2 &	\textbf{65.62} &	97.12 &	\textbf{36.66} &	0.5522\\  
\bottomrule
\end{tabular}}
\label{tab:MRD800}
\end{table}

\begin{figure*}[!htbp]
\centering
\begin{subfigure}{.22\columnwidth}
\centering
\includegraphics[clip=true, trim = 0 0 0 0, width=0.98\linewidth]{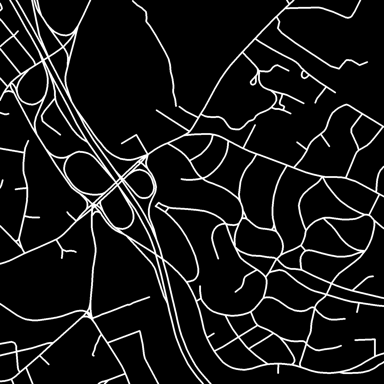} 
\caption{}
\end{subfigure}%
\begin{subfigure}{.22\columnwidth}
\centering
\includegraphics[clip=true, trim = 0 0 0 0, width=0.98\linewidth]{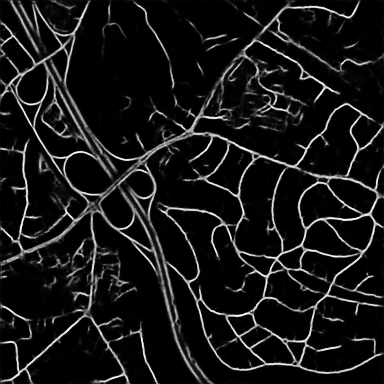} 
\caption{}
\end{subfigure}%
\begin{subfigure}{.22\columnwidth}
\centering
\includegraphics[clip=true, trim = 0 0 0 0, width=0.98\linewidth]{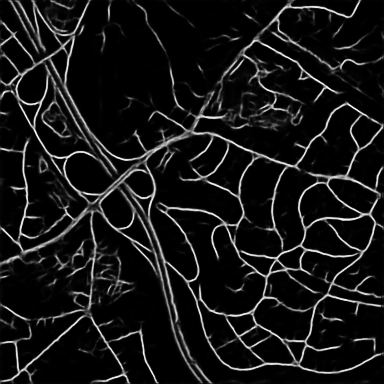} 
\caption{}
\end{subfigure}%
\begin{subfigure}{.22\columnwidth}
\centering
\includegraphics[clip=true, trim = 0 0 0 0, width=0.98\linewidth]{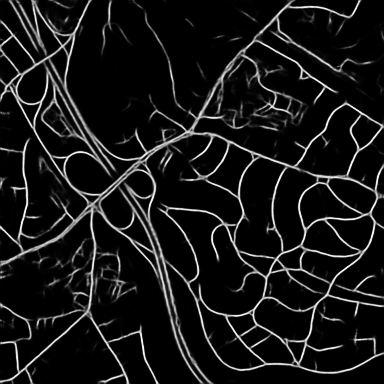} 
\caption{}
\end{subfigure}%
\begin{subfigure}{.22\columnwidth}
\centering
\includegraphics[clip=true, trim = 0 0 0 0, width=0.98\linewidth]{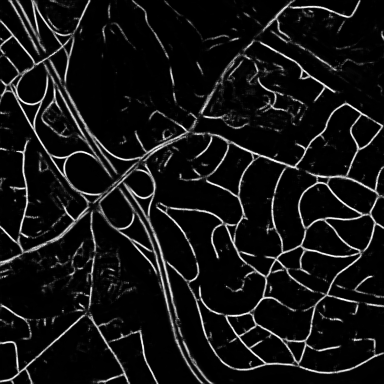} 
\caption{}
\end{subfigure}%
\begin{subfigure}{.22\columnwidth}
\centering
\includegraphics[clip=true, trim = 0 0 0 0, width=0.98\linewidth]{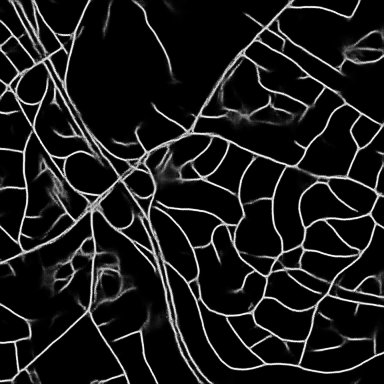} 
\caption{}
\end{subfigure}%
\begin{subfigure}{.22\columnwidth}
\centering
\includegraphics[clip=true, trim = 0 0 0 0, width=0.98\linewidth]{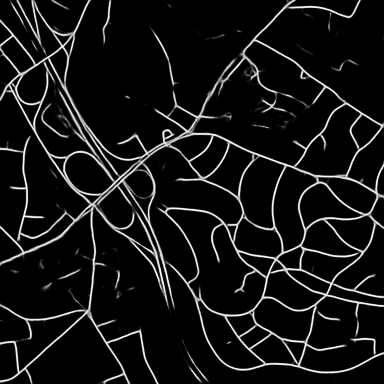} 
\caption{}
\end{subfigure}%
\begin{subfigure}{.22\columnwidth}
\centering
\includegraphics[clip=true, trim = 0 0 0 0, width=0.98\linewidth]{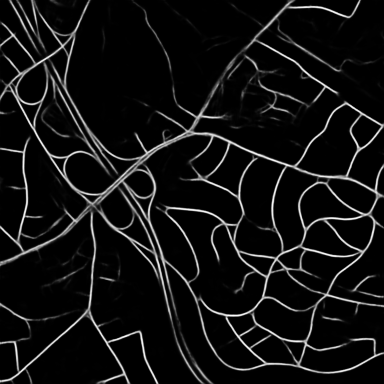} 
\caption{}
\end{subfigure}%
\begin{subfigure}{.22\columnwidth}
\centering
\includegraphics[clip=true, trim = 0 0 0 0, width=0.98\linewidth]{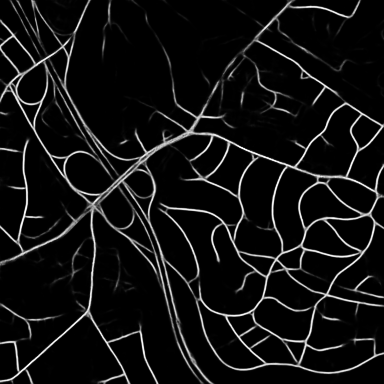} 
\caption{}
\end{subfigure}%

\caption{Segmentation maps over MRD800 dataset using (a) Ground Truth, (b) UNet, (c) UNet++, (d) UNet+++, (e) Attention UNet, (f) SwinUNet, (g) ResUNet, (h) ResUNetFormer-V1, and (i) ResUNetFormer-V2.}
\label{fig:MRD800}
\end{figure*}

In scenario 2, as seen in Table~\ref{tab:MRD800} and Fig.~\ref{fig:MRD800}, statistical result analysis and visual interpretation illustrated the superiority of the ResUNetFormer-V2 with the HetConv operations over the other vision-based algorithms using a recall, dice coefficient, F-1 score, and precision of 36.66\%, 0.5522, 66.62\%, and 97.12\%, respectively. The ResUNet segmentation algorithm obtained the highest dice coefficient (0.5706). The ResUNetFormer-V2 results enhanced the ResUNet model's precision, F-1 score, and recall by around 5\%, 13\%, and 13\%, respectively.

\begin{table}[!htbp]
\centering
\caption{Segmentation results in terms of F-1 score, Precision, Recall, and Dice coefficient for MRD100IOU dataset.}
\resizebox{0.95\linewidth}{!}{
\begin{tabular}{|c|c|c|c|c|} \toprule
Algorithm &	F-1$\times$100 $\uparrow$ &	Precision$\times$100 $\uparrow$ &	Recall$\times$100 $\uparrow$ & Dice coefficient $\uparrow$ \\
 \midrule
UNet \cite{ronneberger2015u} &	52.64 &	81.46 &	35.94 &	0.4929\\ 
UNet++\cite{8932614}	& 54.36 &	73.05 &	46.62 &	0.5142\\ 
UNet+++ \cite{huang2020unet} &	56.63 &	73.87 &	51.44 &	0.53\\ 
SwinUNet \cite{cao2021swin} &	50.74&	78.02&	37.33&	0.4808\\  
ResUNet \cite{8309343}& 64.57 &	92.23 &	48.08 &	0.6227 \\ 
ResUNetFormer\_V1 &	65.19 &	90.75 &	49.48 &	0.6238\\ 
ResUNetFormer\_V2 &	\textbf{65.3} &	\textbf{92.52} &	 	\textbf{53.19} &	\textbf{0.6269}\\  
\bottomrule
\end{tabular}}
\label{tab:MRD100IOU}
\end{table}

\begin{figure*}[!htbp]
\centering
\begin{subfigure}{.22\columnwidth}
\centering
\includegraphics[clip=true, trim = 0 0 0 0, width=0.98\linewidth]{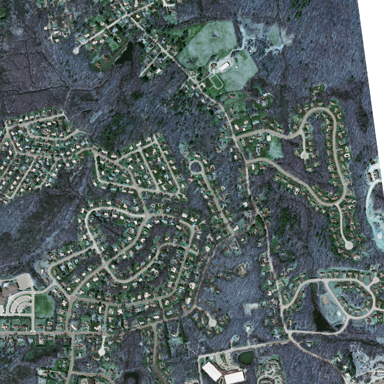} 
\caption{}
\end{subfigure}%
\begin{subfigure}{.22\columnwidth}
\centering
\includegraphics[clip=true, trim = 0 0 0 0, width=0.98\linewidth]{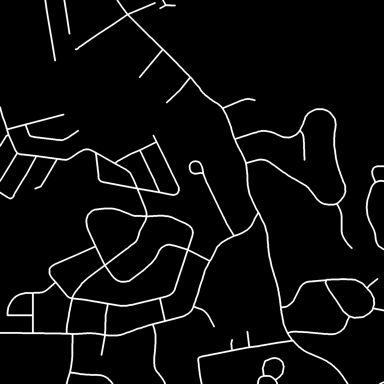} 
\caption{}
\end{subfigure}%
\begin{subfigure}{.22\columnwidth}
\centering
\includegraphics[clip=true, trim = 0 0 0 0, width=0.98\linewidth]{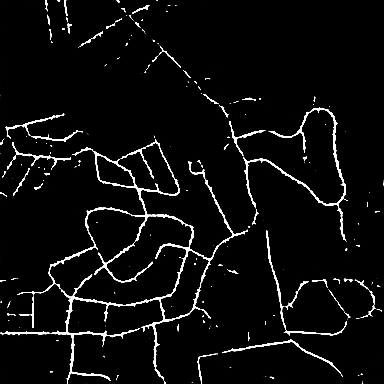} 
\caption{}
\end{subfigure}%
\begin{subfigure}{.22\columnwidth}
\centering
\includegraphics[clip=true, trim = 0 0 0 0, width=0.98\linewidth]{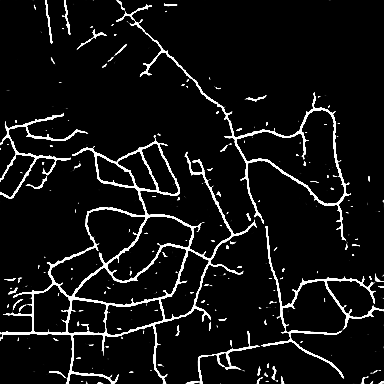} 
\caption{}
\end{subfigure}%
\begin{subfigure}{.22\columnwidth}
\centering
\includegraphics[clip=true, trim = 0 0 0 0, width=0.98\linewidth]{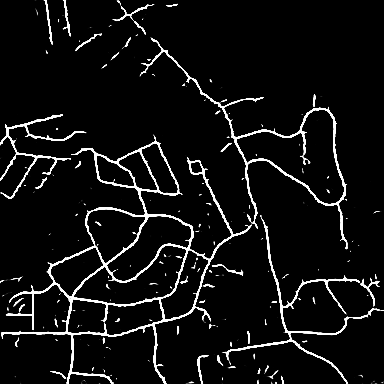} 
\caption{}
\end{subfigure}%
\begin{subfigure}{.22\columnwidth}
\centering
\includegraphics[clip=true, trim = 0 0 0 0, width=0.98\linewidth]{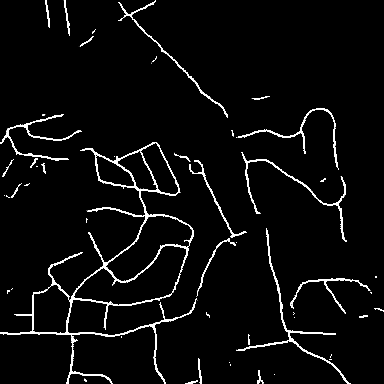} 
\caption{}
\end{subfigure}%
\begin{subfigure}{.22\columnwidth}
\centering
\includegraphics[clip=true, trim = 0 0 0 0, width=0.98\linewidth]{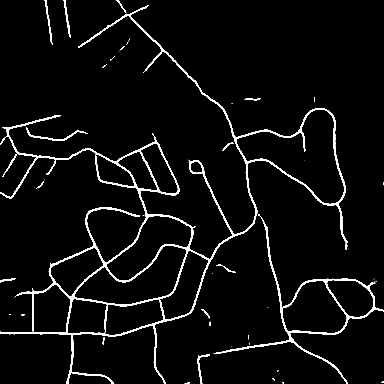} 
\caption{}
\end{subfigure}%
\begin{subfigure}{.22\columnwidth}
\centering
\includegraphics[clip=true, trim = 0 0 0 0, width=0.98\linewidth]{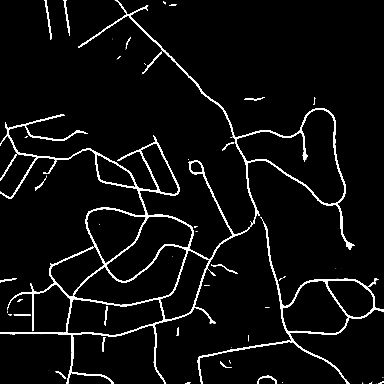} 
\caption{}
\end{subfigure}%
\begin{subfigure}{.22\columnwidth}
\centering
\includegraphics[clip=true, trim = 0 0 0 0, width=0.98\linewidth]{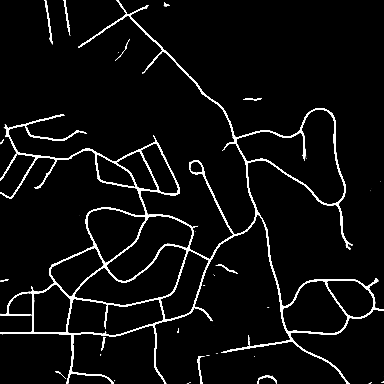} 
\caption{}
\end{subfigure}%
\caption{Segmentation maps over MRD100IOU dataset using (a) RGB image, (b) Ground Truth, (c) UNet, (d) UNet++, (e) UNet+++, (f) SwinUNet, (g) ResUNet, (h) ResUNetFormer-V1, and (i) ResUNetFormer-V2.}
\label{fig:MRD100IOU}
\end{figure*}

In scenario 1 with IoU, statistical analysis and visual interpretation showed better segmentation capability of the ResUNetFormer-V2 over the other semantic segmentation models that achieved a recall, dice coefficient, F-1 score, and precision of 53.19\%, 0.6269, 65.53\%, and 92.52\%, respectively, as reported in Table.~\ref{tab:MRD100IOU} and Fig.~\ref{fig:MRD100IOU}. The ResUNetFormer-V2 enhanced the semantic segmentation performance of the ResUNet  by approximately 1\%, 1\%, 1\%, and 10\%, in terms of F-1 score, precision, dice coefficient, and recall, respectively.

\begin{table}[!htbp]
\centering
\caption{Segmentation results in terms of F-1 score, Precision, Recall, and Dice coefficient for MRD800IOU dataset.}
\resizebox{0.95\linewidth}{!}{
\begin{tabular}{|c|c|c|c|c|} \toprule
Algorithm &	F-1$\times$100 $\uparrow$ &	Precision$\times$100 $\uparrow$ &	Recall$\times$100 $\uparrow$ & Dice coefficient $\uparrow$ \\
 \midrule
UNet \cite{ronneberger2015u} &	53.28 &	80.03	 &39.78 &	0.6296\\ 
UNet++ \cite{8932614}	& 53.66	&78.19	&43.29 &	0.6278\\ 
UNet+++ \cite{huang2020unet}  &	53.99	&71.58&	49.87&	0.5992\\ 
SwinUNet \cite{cao2021swin}  &	50.25	&87.57&	34.11	& 0.6273\\  
ResUNet \cite{8309343} & 58.37&	90&	42&	0.6833 \\ 
ResUNetFormer\_V1 &		\textbf{67.94} &	94.94 &	\textbf{50.33}&	\textbf{0.6860}\\ 
ResUNetFormer\_V2 &	65.62&	\textbf{96.02}	& 46.10&	0.6602 \\  
\bottomrule
\end{tabular}}
\label{tab:MRD800IOU}
\end{table}

\begin{figure*}[!htbp]
\centering
\begin{subfigure}{.22\columnwidth}
\centering
\includegraphics[clip=true, trim = 0 0 0 0, width=0.98\linewidth]{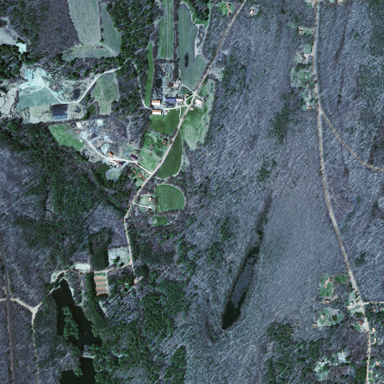} 
\caption{}
\end{subfigure}%
\begin{subfigure}{.22\columnwidth}
\centering
\includegraphics[clip=true, trim = 0 0 0 0, width=0.98\linewidth]{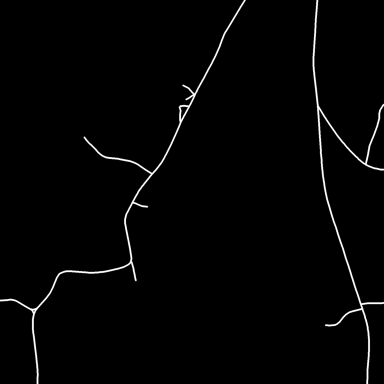} 
\caption{}
\end{subfigure}%
\begin{subfigure}{.22\columnwidth}
\centering
\includegraphics[clip=true, trim = 0 0 0 0, width=0.98\linewidth]{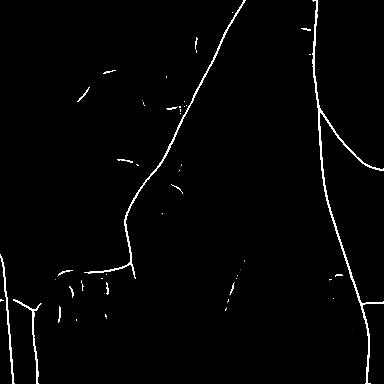} 
\caption{}
\end{subfigure}%
\begin{subfigure}{.22\columnwidth}
\centering
\includegraphics[clip=true, trim = 0 0 0 0, width=0.98\linewidth]{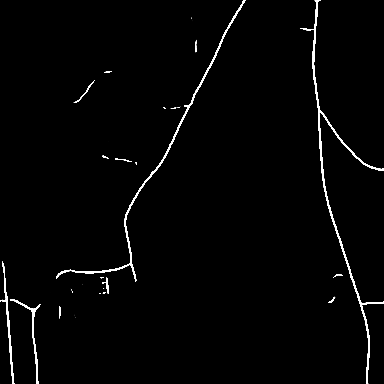} 
\caption{}
\end{subfigure}%
\begin{subfigure}{.22\columnwidth}
\centering
\includegraphics[clip=true, trim = 0 0 0 0, width=0.98\linewidth]{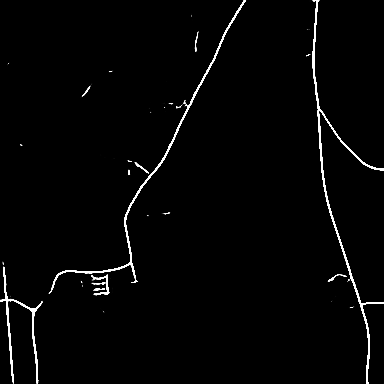} 
\caption{}
\end{subfigure}%
\begin{subfigure}{.22\columnwidth}
\centering
\includegraphics[clip=true, trim = 0 0 0 0, width=0.98\linewidth]{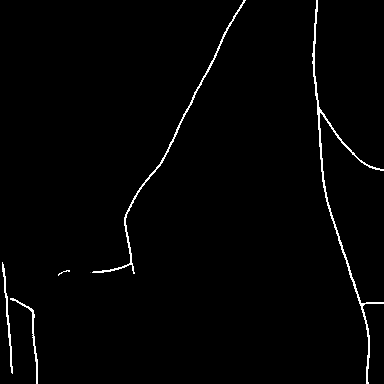} 
\caption{}
\end{subfigure}%
\begin{subfigure}{.22\columnwidth}
\centering
\includegraphics[clip=true, trim = 0 0 0 0, width=0.98\linewidth]{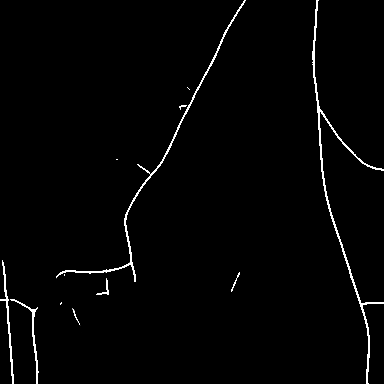} 
\caption{}
\end{subfigure}%
\begin{subfigure}{.22\columnwidth}
\centering
\includegraphics[clip=true, trim = 0 0 0 0, width=0.98\linewidth]{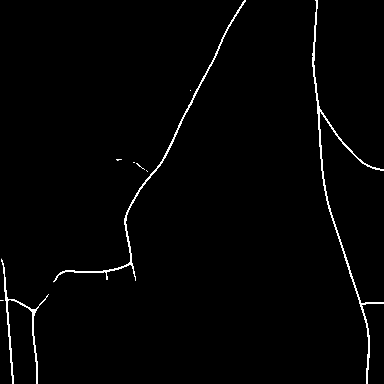} 
\caption{}
\end{subfigure}%
\begin{subfigure}{.22\columnwidth}
\centering
\includegraphics[clip=true, trim = 0 0 0 0, width=0.98\linewidth]{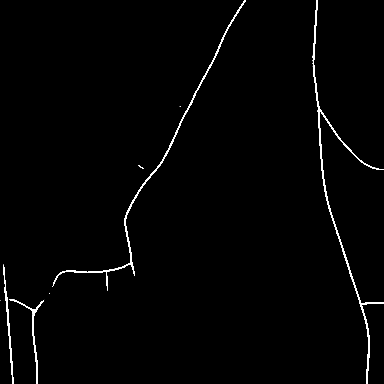} 
\caption{}
\end{subfigure}%
\caption{Segmentation maps over MRD800IOU dataset using (a) RGB image, (b) Ground Truth, (c) UNet, (d) UNet++, (e) UNet+++, (f) SwinUNet, (g) ResUNet, (h) ResUNetFormer-V1, and (i) ResUNetFormer-V2.}
\label{fig:MRD800IOU}
\end{figure*}

\begin{figure*}[!ht]
\centering
\begin{subfigure}{.5\columnwidth}
\centering
\includegraphics[clip=true, trim = 0 75 100 0, width=0.99\linewidth]{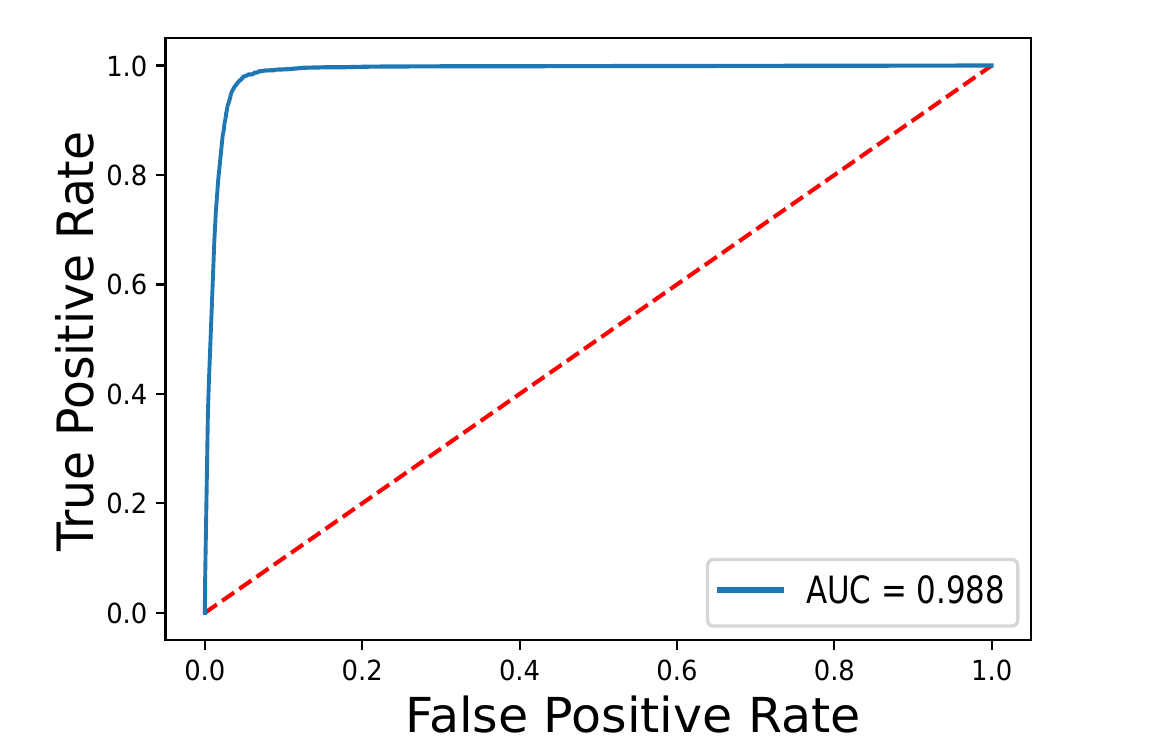} 
\caption{}
\end{subfigure}%
\begin{subfigure}{.5\columnwidth}
\centering
\includegraphics[clip=true, trim = 0 75 100 0, width=0.99\linewidth]{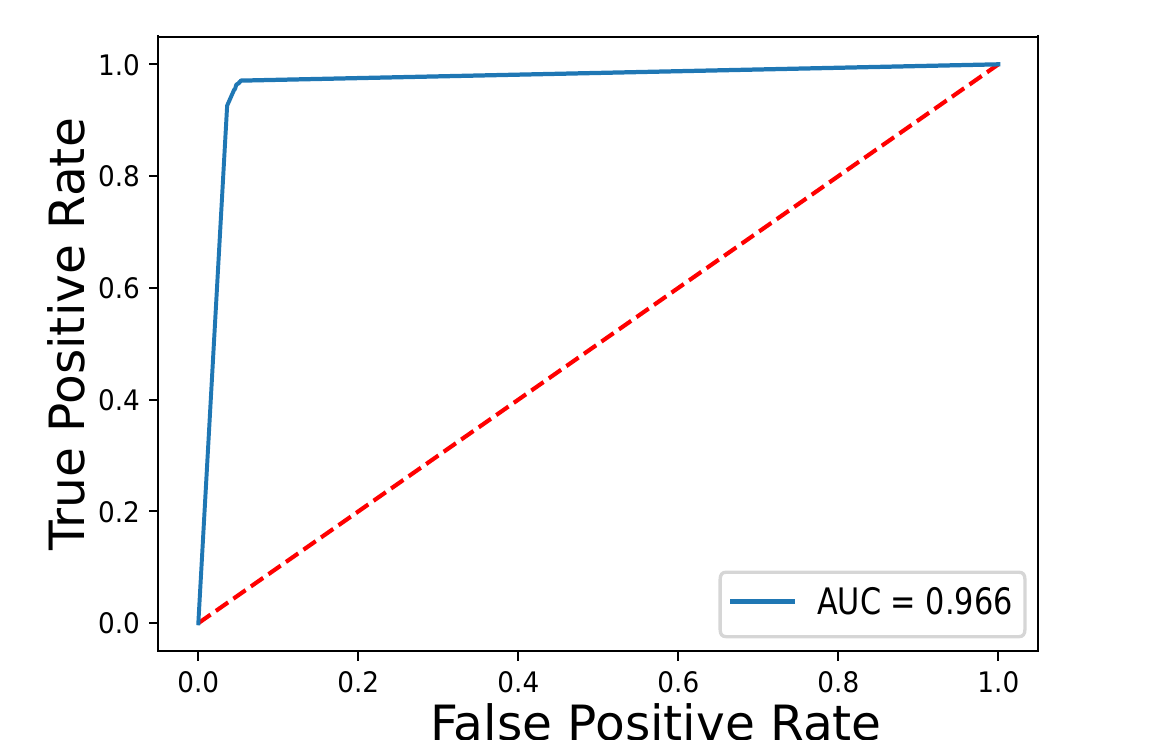} 
\caption{}
\end{subfigure}%
\begin{subfigure}{.5\columnwidth}
\centering
\includegraphics[clip=true, trim = 0 75 100 0, width=0.99\linewidth]{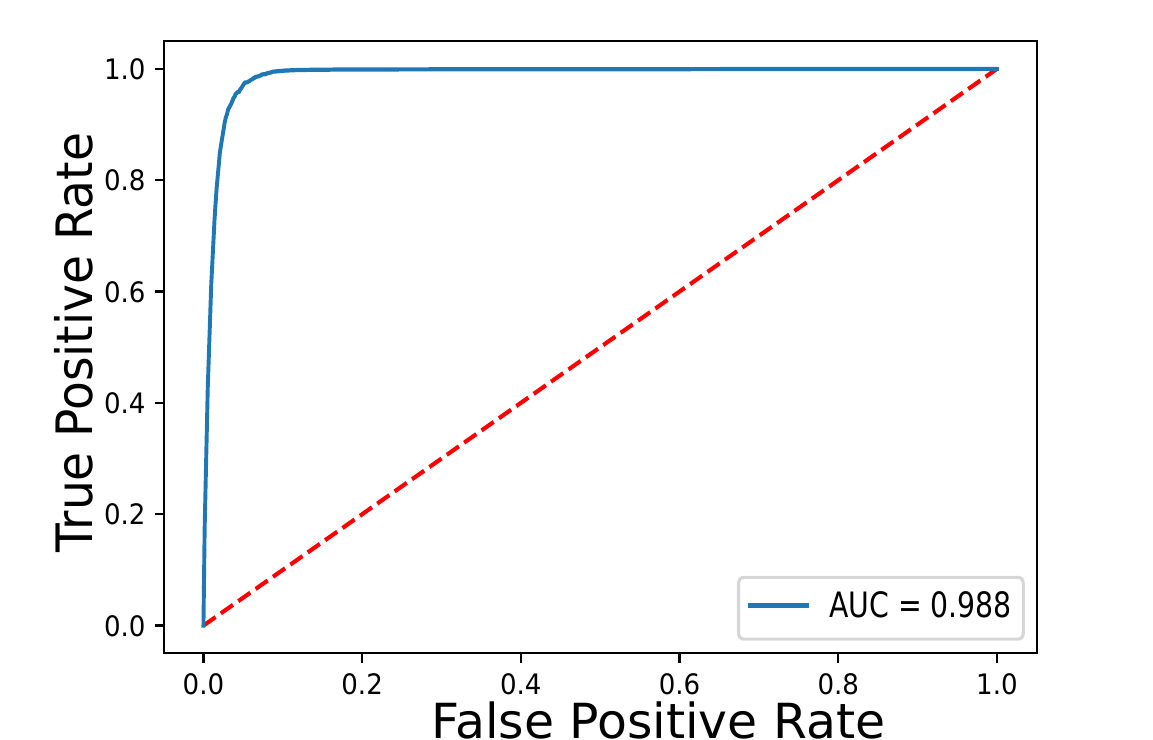} 
\caption{}
\end{subfigure}
\begin{subfigure}{.5\columnwidth}
\centering
\includegraphics[clip=true, trim = 0 75 90 0, width=0.99\linewidth]{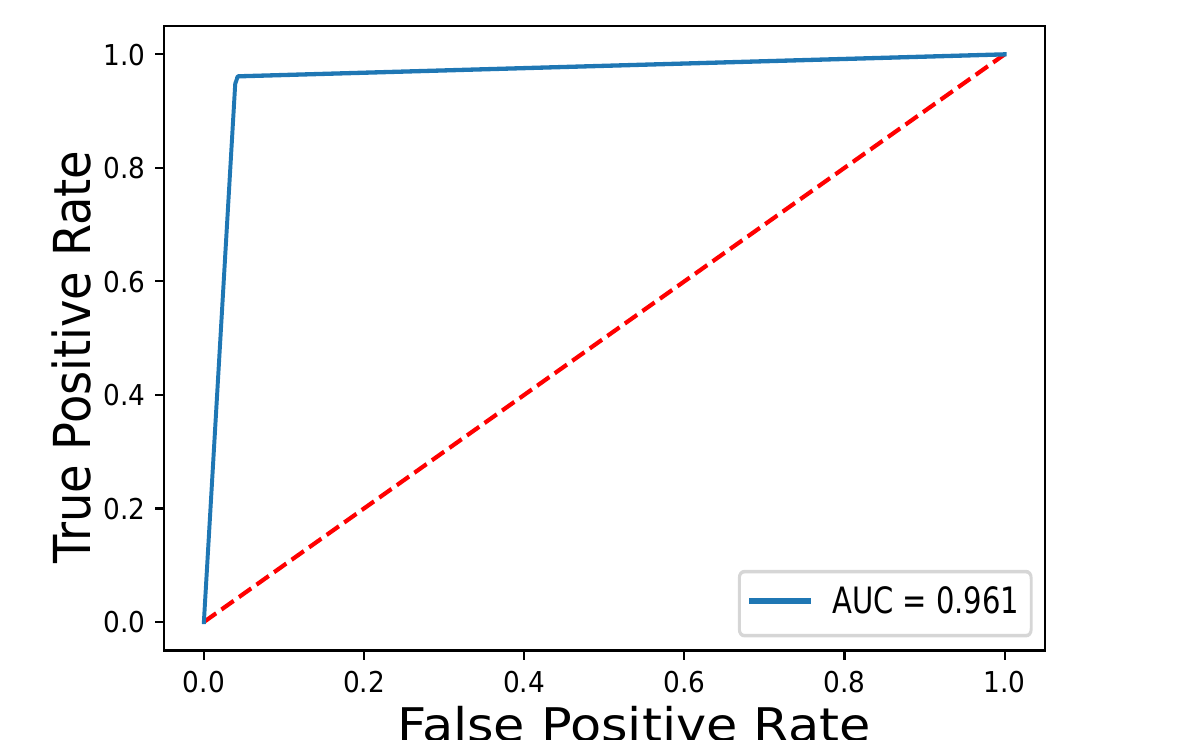} 
\caption{}
\end{subfigure}
\\
\caption{The area under the ROC Curve of the ResUNetFormer-V2 for dataset of (a) MRD100, (b) MRD100IOU, and (c) MRD800, and (d) MRD800IOU.}
\label{fig:AUC}
\end{figure*}

In scenario 2 with IoU, as seen in Table~\ref{tab:MRD800IOU} and Fig.~\ref{fig:MRD800IOU}, the ResUNetFormer-V1 shows superior performance over other semantic segmentation models that obtained a recall, F-1 score, and dice coefficient of 50.33\%, 67.94\%, and 0.686\%, respectively. The ResUNetFormer-V2 algorithm also achieved the highest precision (96.02\%). The results of the ResUNet in terms of dice coefficient, precision, F-1 score, and recall were improved by the ResUNetFormer-V1 by approximately 1\%, 5\%, 14\%, and 17\%, respectively. Moreover, as shown in Fig.~\ref{fig:AUC}, the results demonstrated high Area under the ROC Curve~(AUC) values of 0.988, 0.988, 0.966, and 0.961 for MRD100, MRD800, MRD100IOU, and MRD800IOU, respectively. Results illustrated that the proposed ResUNetFormer-V1 and ResUNetFormer-V2 with the use of NAT led to much better segmentation accuracy and produces much less noisy road maps as compared with the vision and attention-based models like SwinUNet and Attention UNet. Overall, the utilization of the HetConv operations over the standard Conv2D resulted in a better segmentation accuracy.

\section{Conclusion}
\label{sec:con}
This letter proposes and discussed a deep vision transformer-based technique for semantic segmentation, which employs NAT to enhance feature extraction capabilities locally whereas significantly lowering computation costs. The results on the Massachusetts road data demonstrate that the developed model, ResUNetFormer, outperforms statistically and visually the state-of-the-art semantic segmentation models, including UNet, UNet++, UNet+++, Attention UNet, SwinUNet, and ResUNet. ResUNet with HetConv and local attention mechanism operations resulted in much lower noise than that of current CNN and transformer-based semantic segmentation techniques.

\bibliographystyle{IEEEtran}
\bibliography{reference}

\begin{thebibliography}{10}
\providecommand{\url}[1]{#1}
\csname url@samestyle\endcsname
\providecommand{\newblock}{\relax}
\providecommand{\bibinfo}[2]{#2}
\providecommand{\BIBentrySTDinterwordspacing}{\spaceskip=0pt\relax}
\providecommand{\BIBentryALTinterwordstretchfactor}{4}
\providecommand{\BIBentryALTinterwordspacing}{\spaceskip=\fontdimen2\font plus
\BIBentryALTinterwordstretchfactor\fontdimen3\font minus
  \fontdimen4\font\relax}
\providecommand{\BIBforeignlanguage}[2]{{%
\expandafter\ifx\csname l@#1\endcsname\relax
\typeout{** WARNING: IEEEtran.bst: No hyphenation pattern has been}%
\typeout{** loaded for the language `#1'. Using the pattern for}%
\typeout{** the default language instead.}%
\else
\language=\csname l@#1\endcsname
\fi
#2}}
\providecommand{\BIBdecl}{\relax}
\BIBdecl

\bibitem{8309343}
Z.~Zhang, Q.~Liu, and Y.~Wang, ``Road extraction by deep residual u-net,''
  \emph{IEEE Geoscience and Remote Sensing Letters}, vol.~15, no.~5, pp.
  749--753, 2018.

\bibitem{8792386}
X.~Lu, Y.~Zhong, Z.~Zheng, Y.~Liu, J.~Zhao, A.~Ma, and J.~Yang, ``Multi-scale
  and multi-task deep learning framework for automatic road extraction,''
  \emph{IEEE Transactions on Geoscience and Remote Sensing}, vol.~57, no.~11,
  pp. 9362--9377, 2019.

\bibitem{8883072}
Q.~Zou, H.~Jiang, Q.~Dai, Y.~Yue, L.~Chen, and Q.~Wang, ``Robust lane detection
  from continuous driving scenes using deep neural networks,'' \emph{IEEE
  Transactions on Vehicular Technology}, vol.~69, no.~1, pp. 41--54, 2020.

\bibitem{9627165}
D.~Hong, Z.~Han, J.~Yao, L.~Gao, B.~Zhang, A.~Plaza, and J.~Chanussot,
  ``Spectralformer: Rethinking hyperspectral image classification with
  transformers,'' \emph{IEEE Transactions on Geoscience and Remote Sensing},
  vol.~60, pp. 1--15, 2022.

\bibitem{jamali2022deep}
A.~Jamali, M.~Mahdianpari, F.~Mohammadimanesh, and S.~Homayouni, ``A deep
  learning framework based on generative adversarial networks and vision
  transformer for complex wetland classification using limited training
  samples,'' \emph{International Journal of Applied Earth Observation and
  Geoinformation}, vol. 115, p. 103095, 2022.

\bibitem{he2016deep}
K.~He, X.~Zhang, S.~Ren, and J.~Sun, ``Deep residual learning for image
  recognition,'' in \emph{Proceedings of the IEEE conference on computer vision
  and pattern recognition}, 2016, pp. 770--778.

\bibitem{ronneberger2015u}
O.~Ronneberger, P.~Fischer, and T.~Brox, ``U-net: Convolutional networks for
  biomedical image segmentation,'' in \emph{International Conference on Medical
  image computing and computer-assisted intervention}.\hskip 1em plus 0.5em
  minus 0.4em\relax Springer, 2015, pp. 234--241.

\bibitem{long2015fully}
J.~Long, E.~Shelhamer, and T.~Darrell, ``Fully convolutional networks for
  semantic segmentation,'' in \emph{Proceedings of the IEEE conference on
  computer vision and pattern recognition}, 2015, pp. 3431--3440.

\bibitem{dosovitskiy2020image}
A.~Dosovitskiy, L.~Beyer, A.~Kolesnikov, D.~Weissenborn, X.~Zhai,
  T.~Unterthiner, M.~Dehghani, M.~Minderer, G.~Heigold, S.~Gelly \emph{et~al.},
  ``An image is worth 16x16 words: Transformers for image recognition at
  scale,'' \emph{arXiv preprint arXiv:2010.11929}, 2020.

\bibitem{Hassani2022}
\BIBentryALTinterwordspacing
A.~Hassani, S.~Walton, J.~Li, S.~Li, and H.~Shi, ``Neighborhood attention
  transformer,'' 2022. [Online]. Available:
  \url{https://arxiv.org/abs/2204.07143}
\BIBentrySTDinterwordspacing

\bibitem{cao2021swin}
H.~Cao, Y.~Wang, J.~Chen, D.~Jiang, X.~Zhang, Q.~Tian, and M.~Wang,
  ``Swin-unet: Unet-like pure transformer for medical image segmentation,''
  \emph{arXiv preprint arXiv:2105.05537}, 2021.

\bibitem{kingma2014adam}
D.~P. Kingma and J.~Ba, ``Adam: A method for stochastic optimization,''
  \emph{arXiv preprint arXiv:1412.6980}, 2014.

\bibitem{8932614}
Z.~Zhou, M.~M.~R. Siddiquee, N.~Tajbakhsh, and J.~Liang, ``Unet++: Redesigning
  skip connections to exploit multiscale features in image segmentation,''
  \emph{IEEE Transactions on Medical Imaging}, vol.~39, no.~6, pp. 1856--1867,
  2020.

\bibitem{huang2020unet}
H.~Huang, L.~Lin, R.~Tong, H.~Hu, Q.~Zhang, Y.~Iwamoto, X.~Han, Y.-W. Chen, and
  J.~Wu, ``Unet 3+: A full-scale connected unet for medical image
  segmentation,'' in \emph{ICASSP 2020-2020 IEEE International Conference on
  Acoustics, Speech and Signal Processing (ICASSP)}.\hskip 1em plus 0.5em minus
  0.4em\relax IEEE, 2020, pp. 1055--1059.

\bibitem{oktay2018attention}
O.~Oktay, J.~Schlemper, L.~L. Folgoc, M.~Lee, M.~Heinrich, K.~Misawa, K.~Mori,
  S.~McDonagh, N.~Y. Hammerla, B.~Kainz \emph{et~al.}, ``Attention u-net:
  Learning where to look for the pancreas,'' \emph{arXiv preprint
  arXiv:1804.03999}, 2018.

\bibitem{mnih2010learning}
V.~Mnih and G.~E. Hinton, ``Learning to detect roads in high-resolution aerial
  images,'' in \emph{European conference on computer vision}.\hskip 1em plus
  0.5em minus 0.4em\relax Springer, 2010, pp. 210--223.

\end{thebibliography}

\end{document}